\pdfoutput=1
\documentclass[11pt]{article}

\usepackage{acl}
\usepackage{times}
\usepackage{latexsym}
\usepackage{graphicx}

\usepackage[T1]{fontenc}
\usepackage[utf8]{inputenc}

\usepackage{microtype}

\title{Fine-Grained Analysis of Team Collaborative Dialogue}

\author{Ian Perera \and Matthew Johnson \and Carson Wilber \\
Florida Institute for Human and Machine Cognition \\
40 South Alcaniz St., Pensacola, FL 32502 \\
  \texttt{\{iperera, mjohnson, cwilber\}@ihmc.org} \\}

\begin{document}
\maketitle
\begin{abstract}
Natural language analysis of human collaborative chat dialogues is an understudied domain with many unique challenges: a large number of dialogue act labels, underspecified and dynamic tasks, interleaved topics, and long-range contextual dependence. While prior work has studied broad metrics of team dialogue and associated performance using methods such as LSA, there has been little effort in generating fine-grained descriptions of team dynamics and individual performance from dialogue. We describe initial work towards developing an explainable analytics tool in the software development domain using Slack chats mined from our organization, including generation of a novel, hierarchical labeling scheme; design of descriptive metrics based on the frequency of occurrence of dialogue acts; and initial results using a transformer + CRF architecture to incorporate long-range context.
\end{abstract}

\section{Introduction}

Communication between team members is a core component of nearly any task involving multiple people and essential to effective teamwork. Poor communication can lead to disastrous results even in situations where the individuals are competent in their own roles, while exemplary communication can enable performance beyond the sum of individual capabilities. Automatic, fine-grained analysis of team dialogues could potentially identify successes and failures in communication, while identifying productive paradigms, roles, and areas for improvement in collaboration.

Currently, evaluation of team communication during collaborative tasks often depends on human analysis, or if automated, does not typically have a means for explainable analysis of the underlying mechanisms of communication and collaboration that are tied to the task. Common methods of productivity analysis or effort accounting suffer from focusing on quantitative metrics that may not accurately reflect progress and productivity at the appropriate level -- frequent commits or lines of code may mean very little in regards to the progress made on a software solution. Given the prevalence of existing libraries and resources, significant contributions to a piece of software may be composed of only dozens of lines of code. Likewise, team management metrics such as frequency of ticket creation or status updates do not necessarily reflect progress without an understanding of the issues faced or the significance of the effort required to resolve them. Analyzing team discussion provides a much greater insight into collaboration and communication efficiency, as discussions between team members provide a more accurate picture into the significant and meaningful challenges and accomplishments of team members. Furthermore, both teams and individuals can be evaluated according to social and management metrics – e.g., managing frustration, giving praise to increase team morale, taking responsibility for issues and tasks, etc.

For analysis and discussion of methods, we consider the software development domain in which team members of different roles and levels of experience coordinate efforts in a single chat channel over days, weeks, or months. Unlike much prior work, the tasks under discussion are often not explicitly defined and dynamically change in response to new information and progress. Currently we do not have an endpoint to measure team performance in the data collected (and final results of software development tasks, especially in a research setting, are affected by many factors beyond team dynamics), and so we focus on extracting metrics from human-designed patterns of positive and negative team dynamics. For example, repeated requests for clarification may indicate that the listener is not sufficiently familiar with a concept to be able to provide assistance, or the speaker is not clearly describing relevant information. 

\section{Prior Work}
Team communication during collaborative tasks has been the subject of prior work but previous methods on evaluation and analysis of team communication have been limited in multiple ways: they either depend on human analysis, or if automated, they do not have a means for explainable analysis of the underlying mechanisms of communication and collaboration that are tied to the task. Prior methods have used LSA combined with the completion of the task as a signal as to identify positive team communication features \citep{Martin2004automatedteam}, but the language measures identified are automatically learned, not tied to specific team-performance measures, and require a task that is either completed or failed. In many domains, collaborative tasks do not often ``fail'', but rather may go over budget, miss deadlines, or be completed with deficiencies in the final product that are only apparent months or years later. 

To provide actionable feedback in the absence of pass/fail information, \citet{gorman2003} developed specific LSA-based metrics such as semantic communication density and frequency of topic shifts to provide more specific analysis, while \citet{Leshed2009Visualising} focused on real-time feedback based on elements of language use to measure elements of collaboration. However, while these efforts generated quantifiable metrics at a broad dialogue level, they did not measure specific interactions (e.g. task creation and assignment, discussion of solutions, resolution of issues, etc.) in the context of tasks or specific collaborative methods.

Others have used different signals to determine team cohesion, such as documentation coherence \citep{Dong2003designteam, Hill2002}, or multimodal indicators such as vocal features \citep{Neubauer2017}. Methods more closely related to this work include analysis of response patterns of topics in group discussions \citep{Kiekel2001AutomatingMO}. There has also been some work in the educational domain applying NLP to student chats to assess learning performance \citep{Shibani2017, TrausanMatu2010}.  However, we are currently unaware of any prior work that aims to generate a rich, sentence-level labeling structure for a collaborative, multiple-human, task-based dialogue.

Classification of sentences in dialogue according to dialogue acts has primarily been motivated by human-computer interactions, and thus is typically focused on a single task at a time, between two participants, and involves a task that is only advanced through natural language interactions \citep{liu-etal-2017-using-context, Chen2018Dialogue}. Work by \citet{anikina-kruijff-korbayova-2019-dialogue} addressed dialogue act classification in multi-participant dialogues with mixed human-robot teams, although with a more restricted dialogue act taxonomy and without a goal of enabling analysis of team collaboration or social aspects of team communication.

\section{Characteristics of Software Development Chat Data}
Software development chat data, such as what can be found in Slack, poses unique challenges compared to prior dialogue and natural language sources. As opposed to two-participant, Switchboard-style \citep{stolcke-etal-2000-dialogue} dialogues, such dialogues include multiple people who may enter or exit the dialogue at different time points. Furthermore, the intended recipient of utterances is often not explicitly specified, with contextual information needed to resolve it. 

With regards to the underlying tasks, Slack-style chat data can contain multiple tasks and goals, often not explicitly stated as such. In a further deviation from common task-related dialogues found in prior human-computer dialogue work \citep{mctear2002}, discussions do not have a clear start and ending state. Both of these characteristics mean that the underlying goal state is only partially connected to the sequence of dialogue utterances, and recognition of the current goal and task states additionally relies on expert technical knowledge from participants, understanding of roles and responsibilities, and awareness of other development planning information and modalities (such as version control repositories, issue trackers, Kanban boards, etc.).

\section{Dialogue Act Labeling}
To support fine-grained analysis of how team members communicate goals, intentions, questions, and information, we developed a set of labels at the level of specificity of dialogue acts \citep{Austin1962, stolcke-etal-2000-dialogue} -- encompassing the desired function or role of a sentence in the dialogue. The scope of our dialogue acts is designed to target what can be determined from the content of the sentence with some technical familiarity while not requiring a full understanding of the project under discussion. More specifically, we designed the annotation scheme with the following guidelines: 

\textbf{\emph{Hierarchical}} This annotation scheme is intended to be more descriptive than what current state-of-the-art methods with limited datasets may be able to reasonably accomplish. However, to maintain utility with current methods, we designed the labels to be hierarchical, such that any child dialogue act could more broadly be described with an ancestor act. This also allows for methods where a system may want to prefer a less specific, yet still useful classification under uncertainty. 

\textbf{\emph{Distinct}} Aside from the nesting of acts from the hierarchical structure, there should be little to no overlap between dialogue acts to aid in annotation and reduce ambiguity. Furthermore, we design the acts such that a single label can capture nearly all of the function of a sentence. Despite these efforts, there are certain cases where it is nearly impossible to fully encompass all of the functions of an utterance. In these cases, our annotation guide includes priority notes -- labels that should be chosen in the case of potentially ambiguous or overlapping meaning. For example, while ``Excellent'' may be marked as \textit{Social-Appreciation} given its positive sentiment, a function of \textit{Acknowledge-Accept} determined through context should take precedence. 

\textbf{\emph{Actionable}} The precision and level of domain-specificity in the proposed annotation serve the purpose of enabling the extraction of team metrics with specific descriptions of team dynamics. This is in contrast to prior dialogue act annotations, which often are meant to enable an automated system to respond appropriately to a single user. 

The full annotation set has 55 total dialogue acts, with examples given in Table \ref{dialogue-act-examples}. This is a comparatively large number of labels compared to recent dialogue act classification work that utilizes argument and template structure for sentence understanding. The large number of labels is necessary for a labeling process without overlaps or multiple labels per sentence. Furthermore, some labels may be distinguished by other factors beyond textual understanding -- for example, an individual's role in a project may mean that utterances semantically appearing to be a \textit{Request} are actually an \textit{Assign}. 

The large number of labels does make it difficult to gather sufficient examples to ensure coverage when working with small datasets, as would be expected with sensitive data found in Slack chats. We thus consider a reduced set of 18 dialogue acts for this work (9 top-level acts + 9 specific acts that inform metrics in our current set of metrics for evaluating team performance). For any given application, higher-level labels can provide a fallback for sparser data or situations where fine-grained distinctions are unnecessary. The number of labels also makes annotator agreement a challenge -- we are continually refining the annotation guidelines to reduce ambiguity and plan to have agreement results in near-future work. The dataset used in this work consists of labels from a single annotator after an initial agreement pass of three annotators and an associated annotation guide revision.

\subsection{Dataset and Annotation Process}
Our dataset was taken from a team of 8 researchers and programmers engaged in a single project and communicating over Slack chat in a single, shared channel. We obtained a total of 4035 sentences over 6 months worth of chat. Slack data to be annotated is collected in batches of 1-3 weeks worth of chat, with each batch presented as a single document dialogue which includes the speaker name and timestamp for each message. This enables annotators to better interpret intended recipients and possible changes in topic (given a long pause between utterances, for example). Emoticons are presented in text form (e.g. :laughing:), although we currently do not extract Slack threads or label reactions to messages. Annotators are not expected to research details about the project under discussion or reference transcripts from other days. The proportion of each label in the dataset is shown in Table \ref{label-proportions}.

\begin{table}
\centering
\begin{tabular}{lcc}
\hline
\textbf{Label} & \textbf{Train} & \textbf{Test/Dev}\\
\hline
Inform &  0.50 & 0.33\\
Query &  0.11 & 0.12\\
Inform-InResponse &  0.10 & 0.15\\
Acknowledge &  0.09 & 0.10\\
Propose & 0.06 & 0.05\\
Social & 0.04 & 0.07\\
Request &  0.03 & 0.02\\
Assign &  0.03 & 0.03\\
Code & 0.01 & 0.06\\
Reject & 0.01 & 0.04\\
Social-Comradery & 0.01 & 0.01\\
Propose-OfferAssistance & 0.004 & 0.01\\
Social-Frustration & 0.004 & 0.01\\
\hline
\end{tabular}
\caption{\label{label-proportions}
Proportion of reduced-set labels in the training set and the combined test/dev set.
}
\end{table}

Given such a dialogue, annotators are instructed to assign labels to each clause/fragment or emoticon. This permits cases where there may be multiple dialogue acts in a single sentence (e.g. ``Thanks for the update, and I'll get on that'' would be labeled as \textit{Social-Appreciation} for the first clause, and \textit{Acknowledge-Accept} for the second). Conversely, to reduce annotation effort, annotators are allowed to label large blocks of text with multiple sentences with a single annotation, as is often found when a participant is providing a long explanation or series of tasks. These sentences are then split using Spacy \citep{spacy}. 

\begin{table*}
\centering
\begin{tabular}{ll}
\hline
\textbf{Label} & \textbf{Example utterance} \\
\hline
Query-Status & ``How are we doing?''\\
~Query-Status-Personal & ``Jim, any status updates?''\\
~Query-Status-Environment & ``Is this on dev or production?''\\
~Query-Status-TaskOrIssue & ``How’s the work on drone update bug going?''\\
Query-Technical & ``What are the callbacks for the graphics calls?''\\
Query-Admin & ``Who should I speak to for access to the repository?''\\
Query-Through-Uncertainty & ``:confused:''\\
Query-For-Clarification & ``Which one?''\\
Assign-Task & ``Tom, can you take a look at that error?''\\
Assign-Admin & ``Let's have a meeting to regroup''\\
Request & ``Can you send me documentation for that API?''\\
~Request-Help & ``Can you help me figure out this issue?''\\
~Request-Attention & ``@matt''\\
~Inform-Status-Personal & ``Checking''\\
~Inform-Status-Environment & ``ok I've pushed''\\
~Inform-Status-TaskOrIssue & ``I'm currently working on that''\\
~~Inform-Status-TaskOrIssue-Progress & ``UI issues should be fixed now''\\
~~Inform-Status-TaskOrIssue-Impediment & ``The port conflict is preventing us from changing it''\\
Inform-NewIssue & ``The latest update in Spacy broke NeuralCoref''\\
Inform-NewIssue-Anticipated & ``Changing that will likely break the Makefile''\\
Inform-Technical & ``That library requires a CUDA GPU''\\
Inform-Admin & ``Tomorrow's meeting is in room 1212''\\
Inform-InResponse & ``That probably won't work''\\
Inform-ExplainRationale & ``That's to ensure the agents are synchronized''\\
Inform-ClaimTask & ``I'll start working on that''\\
Inform-ClaimProblemResponsibility & ``My bad''\\
Propose-Task & ``We should try to improve the NLP part of the pipeline''\\
Propose-PossibleSolution & ``Try running it as admin''\\
Propose-OfferAssistance & ``Want me to take a look at that?''\\
Propose-Admin & ``We should talk with John about this''\\
Acknowledge-Receipt & ``I see it now''\\
Acknowledge-Accept & ``Will do''\\
Acknowledge-Affirm & ``That's correct''\\
Acknowledge-Validated & ``I pulled the commit and tested it on my machine''\\
Reject & ``That's outside the scope of this project unfortunately''\\
~~Reject-Counter-Assign & ``I can't, Jim, can you handle that?''\\
~~Reject-Counter-Inform & ``I don't see that, it works for me.''\\
~~Reject-Counter-Claim & ``I don't want to give up the computing time for that yet''\\
~~Reject-Counter-Propose & ``What if we use this library instead?''\\
Code-Message-Table-Issue & \textit{<an exception or stack trace>}\\
Code-Message-Table-Solution & \textit{<a series of console commands>}\\
Social-Blame-Person & ``I think that was Rob's commit''\\
Social-Backchannel & ``Check out this video I found''\\
Social-Comradery & ``Joey saves the day again!''\\
Social-Appreciation & ``Thanks, that's great''\\
Social-Frustration & ``Ugh, I've been working on this for hours''\\
\hline
\end{tabular}
\caption{\label{dialogue-act-examples}
Example utterances for the specific dialogue act classes in the taxonomy. Some top-level labels have been omitted as most relevant examples would belong in a more specific label.
}
\end{table*}

\subsection{Annotation Validation} To verify that the above guidelines were upheld with actual data, we generated initial validation metrics to determine whether our annotation scheme was hierarchical and distinct. We trained FastText \cite{Joulin2016bag} on the full label set with no prior language model, then extracted the average semantic vector for each label. We then compared the cosine distance between each pair of labels. If our hierarchical structure is consistent, then vectors within a single top-level labeling class would have a smaller distance than those between classes. Applied to our dataset, we found this to be mostly true (Table \ref{label-distances}). The Social tag has a high variance given the range of emotions and topics that can be expressed (from appreciation, to frustration, to blame, to unrelated topics), and much of the variance in the Query category is attributed to the \textit{Query-Through-Uncertainty} which tends to lack many of the Query indicator words/phrases in the rest of the category. 

Measuring distinctness with statistical significance is more difficult given the small dataset. However, we find that the most similar labels, \textit{Inform-Claim-Task} (\textit{claim ownership of a task, typically if not
explicitly given that task})  and \textit{Reject-Counter-Claim} (\textit{deny change in ownership of something}) still have an average cosine distance of .44.

\begin{table}
\centering
\begin{tabular}{lc}
\hline
\textbf{Label} & \textbf{Avg Distance} \\
\hline
Acknowledge & .85 \\
Assign & .67 \\
Code-Message-Table & .83 \\
Inform & .78 \\
Propose & .75 \\
Query & .90 \\
Reject & .81 \\
Request & .77 \\
Social & .91 \\
\textbf{All} & .91 \\
\hline
\end{tabular}
\caption{\label{label-distances}
Each of the top-level categories has a greater similarity within the category than the average similarity between labels - demonstrated by the smaller distances compared to all label-label pairs (``All'').
}
\end{table}
\section{Team Metrics}
\begin{figure}[h]
\includegraphics[width=\linewidth]{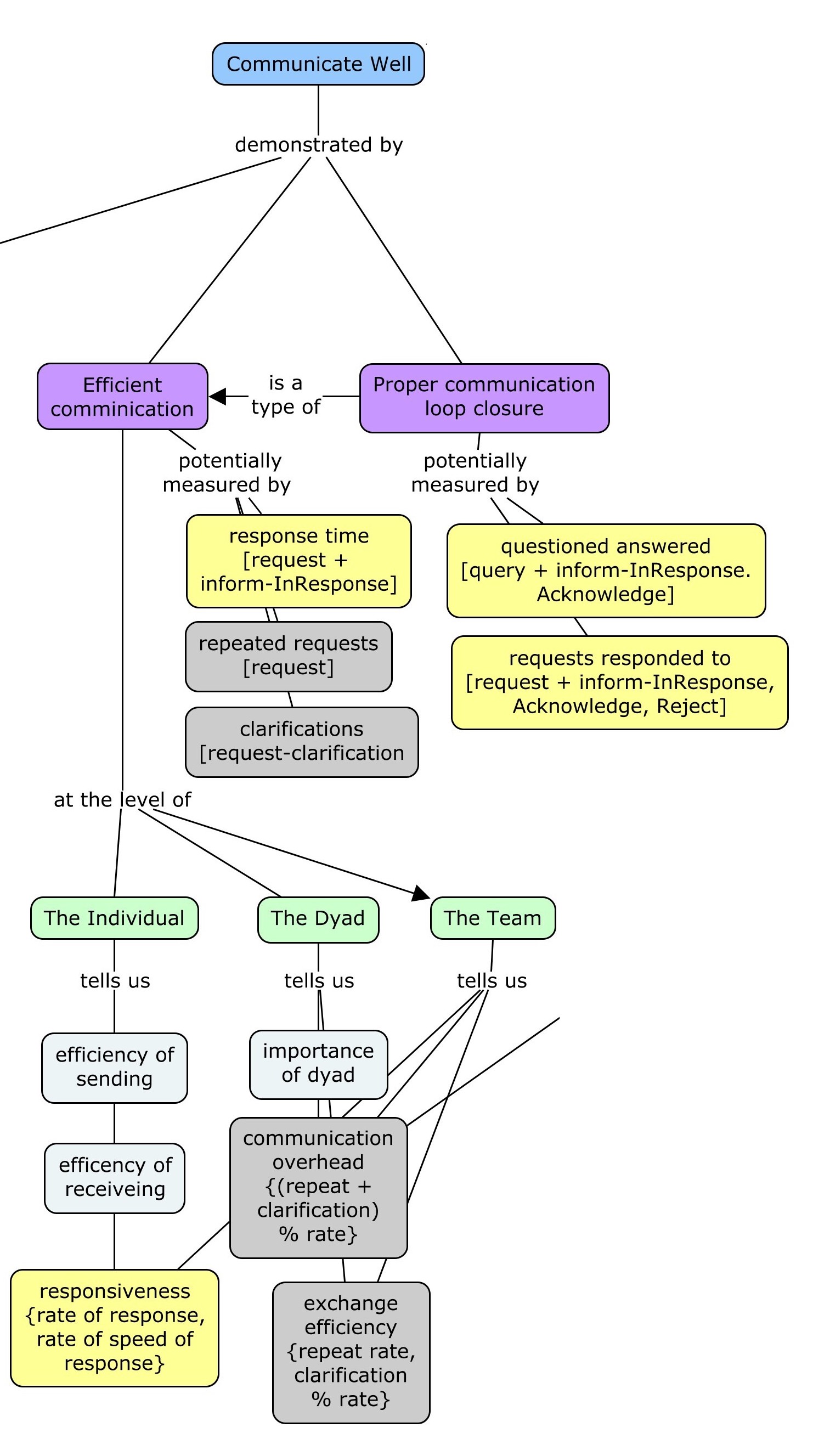}
\caption{\label{team_measures}
An example of how team collaboration analysis is supported by dialogue act classification combined with frequency and dialogue act pair measures. Efficiency of communication is demonstrated by proper loop closure (measured by requests/queries being responded to) and lack of repetition (measured by low frequency of clarification requests).
}
\end{figure}

To provide actionable analysis based on team dynamics, we developed a set of metrics based on signals extracted from the labeled sequence of dialogue acts in a chat stream. These signals could be based on the frequency of certain types of utterances (as in prior work by \citealp{Bowers1998}), or based on the occurrence of statement-response pairs (e.g., is a question followed by an answer), or other quantitative information such as response time or commit frequency (Figure \ref{team_measures}).  Some metrics are nearly universally positive or negative: for example, unanswered queries are always indicative of team communication issues, whereas a reasonable frequency of informative chats and social comradery posts would generally be a positive. However, some of these metrics are not inherently negative or positive, but may provide clues as to potential issues in communication. For example, it may be difficult to distinguish between a communication problem and a conceptually difficult problem requiring significant clarification without the judgment of a knowledgeable human-in-the-loop. 

The guiding principle of these metrics was to develop an assessment of teams according to the following positive team dynamics: communication, coordination, focus on goals, positive collaborative attitude, supportiveness, and adaptability. These principles are consistent with long-standing characteristics of good teamwork \citep{Baker2006Teamwork,Cannon-Bowers1995, Salas2005}. Each of these principles are then evaluated according to one or more metrics based on extracted data from text communication. These metrics focus on specific interactions between team members we extract rather than broad, dialogue-level metrics such as those extracted by \citet{Martin2004automatedteam}. This fine-grained approach enables  analysis of interactions to preemptively identify and study issues and team behavior throughout the course of a project.

\section{Dialogue Act Classification}
To collect metrics on team performance, we must accurately classify dialogue acts in the dialogues described. In this work, we allow for large parts of the dialogue (corresponding to a day of discussion) to be considered as a whole when assigning dialogue acts to utterances. This accommodates the expected use case (i.e., periodic or post-mortem analysis) while not constraining the scope of the work to end-to-end dialogues for projects and tasks that may span months or years. 

Prior work on dialogue act classification has typically focused on domains where the dialogue act of the current utterance can be predicted based on the prior dialogue act, the state of the system, and the text of the utterance. However, given the nature of Slack discussions (multiple interleaved topics with multiple participants), such short-range, singular context is insufficient for aiding in identifying the appropriate dialogue act for an utterance.

Transformer architectures have shown great promise in adapting generic language models to specific tasks with minimal domain-specific data. However, the common BERT \citep{Devlin2018bert} architecture typically operates with context more locally than is needed for this application, and methods that incorporate text at the document level, e.g. Longformer \citep{Beltagy2020longformer}, do not provide a means for fine-grained labeling of sentences. 

\subsection{Sequential Sentence Classification in Slack Chat Data}
To address these limitations, we utilize work on sequential sentence classification by \cite{cohan-etal-2019-pretrained}, where sentences are simultaneously labeled in a sequence of about 8-10 sentences and there is expected contextual information contributed by each of the sentences. In that work, the sequences were sentences within research paper abstracts labeled as BACKGROUND, OBJECTIVE,  METHOD,  RESULT, or OTHER. Utilizing the sentences taken as a sequence allows a model to learn patterns such as the ordering of these sub-topics, whether all of the relevant sub-topics are covered, and that the same label will tend to occur multiple times in adjacent sentences. 

In the Slack chat domain, however, there are significant differences in the contextual assumptions a sentence labeling model can make. First, while we expect common patterns of responses (e.g. multiple sequential \textit{Inform}s, pairs of \mbox{\textit{Query->Inform}}, \mbox{\textit{Assign->Acknowledge}}, \mbox{\textit{Inform->Query}}, etc.)  there is no clearly defined start and end point for where a sequence of sentences should be segmented from the rest of the dialogue. Thus patterns will often occur contained within or intersecting other patterns. Responses may also be delayed by hours or even days, depending on workload and the number of projects a participant is involved in. 

Furthermore, with the goal of identifying deviations from positive team coordination and communication, we must be careful not to overbias the model to expect certain patterns. For example, if our model is expecting positive team communication metrics such as \textit{Query}s being answered, it may miscategorize instances which vitally demonstrate a breakdown in communication. 

\subsection{Experimental Methods}
Because the work by \citeauthor{cohan-etal-2019-pretrained} used a domain-specific BERT model for their highest-performing configuration, we instead used their strong baseline model which combines a transformer with a CRF layer. The CRF layer provides a method for storing the context and likely transitions between dialogue acts without overly biasing the semantic classification layers. 

Because the CRF context layer only operates on a subsequence of the larger dialogue and is limited to a certain length given constraints on the transformer architecture, the choice of an appropriate segmentation for developing a dialogue act model is vital for improving performance. Context across subsequence boundaries is lost, and there is often no explicit indicator that a context shift has occurred. We experiment with several different segmentation methods -- a static window of 5 or 10 utterances, a window based on time elapsed between messages, and segmenting sequences based on speaker changes. We also provide results using FastText without using pretrained embeddings to illustrate the importance of even a general-purpose pretrained language model.

\textbf{\emph{Static Window}} To establish a baseline without dialogue context, we first perform an experiment with the subsequence consisting only of sentences contained within an individual's chat message (\textit{1-line}). The simplest method for incorporating context is to segment dialogue utterances with a static size, in this case either 5 (\textit{5-line}) or 10 (\textit{10-line}) utterances. There is some variance from these numbers as sentences that are split from a single label are treated as multiple utterances but then can exceed the 5 or 10 line count limit. The drawback of this method is that it uses no information about when prior context is relevant, and so the subsequence boundaries can be misaligned with those sentences that are contextually relevant to each other.

\textbf{\emph{Time Limit Window}} By the nature of Slack chats and the urgency of their role in workplace communication (more rapid than email, less rapid than in-person meetings or phone calls), hours may pass before a response is received. However, in times when team members are highly engaged and pushed to coordinate for an upcoming deadline, their utterances are likely to be more rapid and responsive. We hypothesized that utterances more closely occurring in time are more likely to be related, and thus we experimented with a temporal segmentation within the 10-line limit (\textit{10-line-time}). Once an hour had passed, a new dialogue window is created to capture the fact that after a long delay, a new discussion topic is more likely.

\textbf{\emph{Speaker Window}} Finally, we experimented with restricting context to two speakers at a time (\textit{10-line-speaker}). The context window remained open until either the 10 line limit was reached or a new team member (beyond the two speaker limit) posted a message, thus starting a new window and resetting the involved speakers. This assumes that topic discussions only involve two speakers at a time, and thus fails to accurately segment when multiple people are involved in discussing an issue.

We used a train/dev/test split of roughly 80\%/5\%/15\%, although there is some variation as we did not split subsequences. The splits were chosen randomly, with all of the dev data consisting of sequential utterances and the test data consisting of two sequential batches. We use the baseline configuration used by \citeauthor{cohan-etal-2019-pretrained} (RoBERTa pretrained transformer \cite{liu2019roberta}, 0.1 dropout, Adam optimizer with learning rate of 1$e$-5). Training took 2-3 hours on a single consumer-level CPU.

\section{Results}
Results for varying segmentation methods are shown in Table \ref{results}. Because every sentence can be given a label in our annotation scheme, we report accuracy as our metric for performance. FastText scored the lowest of all methods, likely due to the combination of a large number of labels, lack of pretrained embeddings, and limited data.  The lower performance in the \textit{1-line} configuration with a steady increase with higher line counts demonstrates the importance of context for classifying utterances. 

The speaker split performed worse than a static split, likely because many conversations involved multiple participants (Figure \ref{fig:dialogue}). It is not uncommon in software development (and was often the case in this dataset) that 3 or more people may be involved in a single issue -- one for ensuring an idea is carried out according to higher-level goals and intentions, one responsible for the implementation, and one or more providing assistance regarding infrastructure, ideas, or possible ramifications. The time split yielded the highest performance, indicating that for this team, temporal proximity serves as a relatively good indicator of relevance. 
\begin{center}
\fbox{
\begin{minipage}{18em}
\textbf{PG:} BR, don't change rigid body boundaries or origin without letting me know. \{\textit{Request}\} \\
\textbf{BR:} ill check \{\textit{Acknowledge}\} \\
\textbf{ER:} Since we are doing distance calculations, please put a table listing the following: location of origin, distance to top of object, distance to bottom of object \{\textit{Assign-Task}\} 

\end{minipage}
}
\captionsetup{width=0.48\textwidth}
\captionof{figure}{An example of multi-participant dialogue that would be improperly segmented assuming only pair discussions.}
\label{fig:dialogue}
\end{center}
\begin{table}
\centering
\begin{tabular}{lc}
\hline
\textbf{Configuration} & \textbf{Test Accuracy} \\
\hline
FastText & .44 \\
1-line & .58 \\
5-line & .59 \\
10-line & .60 \\
10-line-speaker & .58 \\
10-line-time & .61 \\
\hline
\end{tabular}
\caption{\label{results}
Accuracy of different configurations in sequential sentence classification for labeling dialogue acts. \textit{\#-line} indicates the number of sentences considered together as a sequence, with \textit{speaker} indicating splits based on speaker changes, and \textit{time} indicating splits after an hour had passed between messages.
}
\end{table}

\section{Discussion}
While the time-based method led to the best performing configuration in this case, the most effective method is likely team-dependent. Smaller teams may benefit more from speaker-specific context, and teams that are more frequently engaged may benefit from a different type of context (such as topic- or thread-based). In future work we plan to explore how these different methods might be combined into a more flexible and dynamic contextual system.

We believe this work is only the starting point for future efforts in this space. Advancements in contextual representations, methods incorporating hierarchical or embedded labeling schemes, or topic- or semantic-based segmentation of sentence streams could greatly improve performance in this task. Additionally, a more sophisticated representation of tasks, efforts, roles, and capabilities could potentially lead to both a higher performing system and a better understanding of the characteristics and dynamics of collaborative projects.

\section{Ethical Considerations}
In an era of increasingly remote work, especially in the software development field, semi-automated analysis of team communication and collaboration has many potential ethical considerations. A person's contribution to a team cannot be fully captured by a few metrics based on dialogue acts, and a well-functioning team may be able to succeed even with very little communication. Beyond that, certain individuals may be unfairly penalized by an over-reliance on these metrics. For example, someone unversed in online chat social cues may find that their \textit{Social-Comradery} measure is lower, when in fact it is high when considering their in-person interactions. Another individual may find themselves at the center of problem situations simply because they are typically the one who can fix the problems encountered. Finally, responsiveness measures taken alone can easily penalize individuals who attempt to preserve a work-life balance in consideration of long-term mental health and performance, or who simply have other obligations such as children or family that need around-the-clock care. 

On the other hand, such a system also has the potential to highlight positive qualities of individuals that may otherwise be overlooked. Cohesiveness and comradery are often qualities missed in traditional analytics and performance measures, yet would be rewarded in a system that uses our metrics. Toxic behavior such as blaming and directed frustration would be penalized, thus working to emphasize mental health of team members over solely focusing on hitting performance metrics. Individual expressions of frustration could also be an indicator for high-stress situations, which should be monitored for their potential to lead to burnout and poor decision-making.

We emphasize a human-in-the-loop system for analysis because a fully automated system is currently unable to conceptualize a holistic view of individuals and teams. This system can be used most effectively and ethically by leadership that appropriately focuses on long-term positive relationships of team members over easily quantifiable metrics that need to be considered in the context of the team, project, and organization.

\section{Acknowledgements}
This work was funded by ARO Contract number (W911NF20C0007). The views, opinions and/or findings expressed are those of the authors and should not be interpreted as representing the official views or policies of the Department of Defense or the U.S. Government.
\bibliography{anthology,custom}
\bibliographystyle{acl_natbib}

%\appendix

%\section{Example Appendix}
%\label{sec:appendix}

%This is an appendix.

\end{document}